\documentclass{article}

\PassOptionsToPackage{round}{natbib}

\usepackage[preprint]{neurips_2022}

\usepackage[utf8]{inputenc} 
\usepackage[T1]{fontenc}    
\usepackage{hyperref}       
\usepackage{url}            
\usepackage{booktabs}       
\usepackage{amsfonts}       
\usepackage{nicefrac}       
\usepackage{microtype}      
\usepackage{xcolor}         
\usepackage{comment}
\usepackage{graphicx} 
\usepackage{amsmath,amsthm}
\usepackage{color,soul}

\usepackage{tikz}
\usetikzlibrary{bayesnet}
\usetikzlibrary{arrows}
\usepackage{color}
\usepackage{graphicx}
\usepackage{caption}
\usepackage{subcaption}
\usetikzlibrary{backgrounds}

\theoremstyle{plain}

\newtheorem{proposition}{Proposition}

\newcommand{\E}{\mathbb{E}}
\newcommand{\bX}{\mathbf{X}}

\newcommand{\bM}{\mathbf{M}}
\newcommand{\bm}{\mathbf{m}}

\title{Estimating Causal Effects Under Image Confounding Bias with an Application to Poverty in Africa}%

\author{%
  Connor T. Jerzak \\
  Institute for Analytical Sociology\\
  Linköping University \\
  Email: \texttt{connor.jerzak@liu.se} \\
  Website: \texttt{ConnorJerzak.com} \\
   \And
    Fredrik Johansson  \\
   Data Science and AI Division \\
   Chalmers University of Technology \\
   Email: \texttt{fredrik.johansson@chalmers.se} \\
   Website: \texttt{fredjo.com}
   \AND
   Adel Daoud \\
  Institute for Analytical Sociology\\
  Linköping University \\
  Email: \texttt{adel.daoud@liu.se} \\
  Website: \texttt{AdelDaoud.se} \\ 
  AI and Global Development Lab: \texttt{global-lab.ai}
}

\begin{document}
\maketitle

\begin{abstract}
Observational studies of causal effects require adjustment for confounding factors. In the tabular setting, where these factors are well-defined, separate random variables, the effect of confounding is well understood. However, in public policy, ecology, and in medicine, decisions are often made in non-tabular settings, informed by patterns or objects detected in images (e.g., maps, satellite or tomography imagery). Using such imagery for causal inference presents an opportunity because objects in the image may be related to the treatment and outcome of interest. In these cases, we rely on the images to adjust for confounding but observed data do not directly label the existence of the important objects. Motivated by real-world applications, we formalize this challenge, how it can be handled, and what conditions are sufficient to identify and estimate causal effects. We analyze finite-sample performance using simulation experiments, estimating effects using a propensity adjustment algorithm that employs a machine learning model to estimate the image confounding. Our experiments also examine sensitivity to misspecification of the image pattern mechanism. Finally, we use our methodology to estimate the effects of policy interventions on poverty in African communities from satellite imagery. 

\vspace{.10in}
 
\vspace{.05in}
\noindent {\bf Word count: } 7,016

\vspace{.05in}
\noindent {\it Note: } This work is largely subsumed by 
\begin{itemize}
\item[] Jerzak, Connor T., Fredrik Johansson, and Adel Daoud. ``Integrating Earth Observation Data into Causal Inference: Challenges and Opportunities.'' {\it arXiv preprint} \url{arXiv:2301.12985} (2023).
\end{itemize}

\end{abstract}
\newpage

\section{Introduction}\label{ss:intro}
As learning algorithms increasingly get deployed to inform decision making, a growing literature has emerged that seeks to provide valid estimation of treatment effects in high dimensions \citep{li2021bounds,mozer2020,chernozhukov2018double,yoon2018ganite,wager2018estimation,shalit2017estimating,hill2011bayesian,schneeweiss2009high, Belloni2014, alexanderOverlapObservationalStudies2021, daoudStatisticalModelingThree2020}. Yet there is a lack of methodological frameworks for causal estimation when confounding is induced by patterns or objects observed in an image  \citep{castroCausalityMattersMedical2020}. 

Images are prevalent not only in social media, but also in the health setting and humanitarian context. For instance, after analyzing an X-ray image of a cancer patient, a doctor may alter their treatment protocol for this patient; the diagnostic information in the image is also correlated with the survival outcome  \citep{castroCausalityMattersMedical2020}. Likewise, a policymaker may consult maps to evaluate where to allocate aid to villages that otherwise may remain poor \citep{holmgrenSatelliteRemoteSensing1998a,bediMorePrettyPicture2007}. Besides annotated maps---where objects such as forests, roads, and cities are named in the map---policymakers are increasingly reliant on raw satellite images where no annotation exist. For example, since 2000, policymakers frequently rely on raw satellite images to evaluate damage due to natural disasters or war \citep{voigtGlobalTrendsSatellitebased2016, burkeUsingSatelliteImagery2021, kinoScopingReviewUse2021}.  
Based on these images, they decide where to intervene helping the poor  \citep{borieMappingResilienceCity2019,daoudUsingSatellitesArtificial2021a}.

These disparate examples share a common structure: an actor examines an image $I$, looking for certain latent patterns $U$ which guide the choice of intervention $T$. These observed patterns indicate the existence of real-world objects which directly influence the outcome of treatment $Y$, thereby inducing confounding if not adjusted for. However, as image patterns $U$ are often unlabeled, even when $I$ is observed, these patterns and objects are difficult to adjust for directly \citep{voigtGlobalTrendsSatellitebased2016}.

In this article, we study observational causal inference in the presence of confounding due to latent image patterns. We describe several causal structures plausibly consistent with real-world applications and discuss their implications on identification and estimation. We focus on under what conditions the image alone is sufficient to adjust for the confounding introduced by $U$. This holds, for example, when $U$ may be derived deterministically from $I$. An important special case of image pattern confounding occurs when decision makers make choices based on the (translation-invariant) existence of a pattern in the image, which motivates adjustment techniques based on convolutional models. 

We study finite-sample estimation of average treatment effects in the fully identified case by conducting a simulation in which confounders are derived from convolution of a 2D filter with the observed image. In this setting, we investigate the impact of model misspecification on estimates. Finally, we demonstrate the use of the proposed estimation framework in an application in which we evaluate the effect of international aid programs on poverty by estimating treatment propensity in a geographic region with a convolutional neural network, motivated by our identification results. 

There are several important contributions for how algorithms may discover causal structures from images \citep{chalupkaMultiLevelCauseEffectSystems2016,chalupkaUnsupervisedDiscoveryNino2016,chalupkaVisualCausalFeature2015,scholkopfCausalRepresentationLearning2021,yiCLEVRERCoLlisionEvents2020,dingDynamicVisualReasoning2021}, but there is a lack of a systematic analysis how to use images for causal inference, where latent objects in the image may affect treatment and outcome \citep{castroCausalityMattersMedical2020}.  Image data is omnipresent, from online advertisement to precision agriculture.  Thus, developing techniques for \emph{causal inference under image pattern confounding} would open up new avenues for observational studies \citep{pawlowskiDeepStructuralCausal2020, singlaExplanationProgressiveExaggeration2020, kaddour2021causal}.

\section{Related Work and Contribution} 
While the literature on generalized linear regression modeling is well-developed in articulating the bias in model parameter estimation due to spatial dynamics \citep{paciorek2010importance}, there have been fewer works on the nature of image-based confounding in the causal inference context \citep{castroCausalityMattersMedical2020, reddyCANDLEImageDataseta, duAdversarialBalancingbasedRepresentation2021}. Some examples include work on estimating counterfactual outcomes under spatially defined counterfactual treatment strategies  \citep{papadogeorgou2020causal}, on accounting for spatial interdependence in causal effect estimation \citep{reich2021review}, and on balancing covariate representations using adversarial networks and image data  \citep{kallus2020deepmatch}. Other works address images as treatments \citep{kaddour2021causal}, counterfactual inference and interpretability \citep{pawlowskiDeepStructuralCausal2020, singlaExplanationProgressiveExaggeration2020}, and image-based treatment effect heterogeneity \citep{jerzak2022image}.

Our work is related to the literature on identification via proxies~\citep{tchetgen2020introduction} or drivers~\citep{pearl2013linear} of confounders. For the former, \citeauthor{louizosCausalEffectInference2017a}, developed Causal Effect Variational Autoencoder (CEVAE) which uses proxies to infer the distribution of the latent confounder and use this in  adjustment. In contrast, our approach adjusts for an observed variable---the image. Our article formalizes key assumptions required for the correctness of this method, and provided a general framework for conducing causal inference using images, where unlabeled objects in the image may affect both treatment and outcome \citep{castroCausalityMattersMedical2020}. This image-based confounding bias might in some circumstances be equivalent to traditional spatial interdependence, but differs insofar as the confounding bias is defined with reference to unlabeled entities in the image, thereby injecting bias \citep{paciorek2010importance}. Relying on our formalization and model implementation, we analyze  aid interventions (treatment) and poverty (outcome) in Africa. 
As discussed in \S\ref{ss:intro}, policymakers often rely on satellite images for aid intervention \citep{voigtGlobalTrendsSatellitebased2016, bediMorePrettyPicture2007}. Thus, our approach also facilitates future policymaking---hopefully contributing to the use of AI for social good. 



\section{Causal Effect Estimation Under Image Pattern Confounding}\label{ss:theory}

\paragraph{Motivation}
Because each image is defined at a more granular resolution than the unit of analysis, we can use it to potentially reconstruct some of unobserved confounders by learning the function generating confounder from image. For example, assume that researchers seek to analyze the effect of a village-level treatment. From a government census, they obtain mean income information for the village ($s$) and then perform an analysis assuming $Y_s(0),Y_s(1)\perp T_s|X_s$, where $X_s$ contains the income data, $T_s$ is the treatment, and $Y_s(t)$, the potential outcome under $t\in\{0,1\}$. Yet, unless the mean income is the true confounder, the analysis will still be biased, which would occur if, in fact, \emph{minimum} income drove the decision to allocate $T_s$. However, using satellite images for each scene, we seek to reconstruct the minimum income signal based on our access to the higher resolution data. 

In other words, we can weaken the assumption used in many empirical analyses that the variables measured at the scene-level in fact contain the true confounders, when in fact there may be highly non-linear functions which use more granular information in generating the confounding structure. With image data, we, in principle, can hope to reconstruct some of those factors using advances in image-based machine learning models effective in the prediction domain (e.g., \citet{sun2013deep}).

\paragraph{Baseline Confounding Model}
With this motivation in mind, we study identification and estimation of the average treatment effect (ATE) of  $T$ on a real-valued outcome of interest, $Y$, based on observational data. With $Y(1)$ being the potential outcome~\citep{rubin2005causal} under intervention with $T=1$ and $Y(0)$ for $T=0$, 
$$
\text{ATE} = \E[Y(1) - Y(0)]~.
$$
The ATE can represent, for example, the difference in average wealth in villages after anti-poverty interventions and after no intervention, respectively. In our setting, historical interventions were determined by a decision-maker in the context of patterns derived from an image $I$ using an unknown process. Although we focus on a decision-maker being affected by patterns in an image (that is $I \to U$), our approach could be generalized to the case where a phenomena (e.g., an event of natural disaster) is imprinted in an image ($U \to I$) and that phenomena affects the outcome (poverty) and the probability of humanitarian aid (treatment). The former is called a \textit{proxy} and the latter is a \textit{driver}, yet enable the same type analysis \citep{pearlLinearModelsUseful2013} that is found in the proxy literature \citep{kurokiMeasurementBiasEffect2014,louizosCausalEffectInference2017a}. To understand estimation dynamics of the ATE from observational images, and adjust for induced confounding bias, we analyze the data generating process next.

\label{ss:structure}

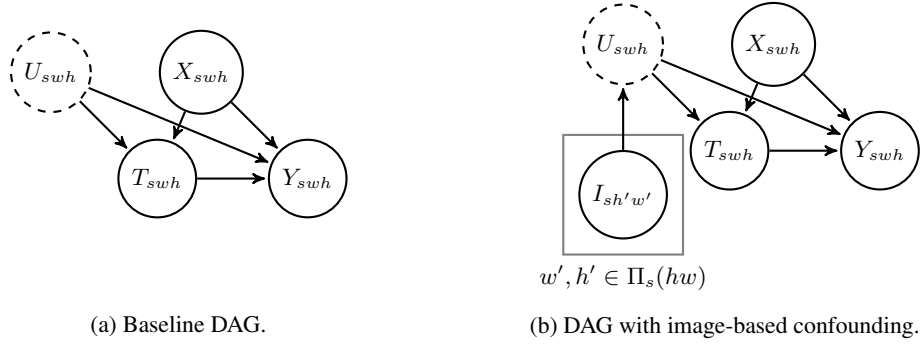
\begin{figure}[t]
    \begin{subfigure}{0.48\textwidth}
        \centering
        \vspace{1em}
        \begin{center}
        \tikzstyle{main node}=[circle,draw,font=\sffamily\small\bfseries]
        \tikzstyle{sub node}=[circle,draw,dashed,font=\sffamily\small\bfseries]
        \begin{tikzpicture}[->,>=stealth',shorten >=1pt,auto,node distance=2cm,thick]
          
          \node[sub node] (1) {$U_{swh}$};
          \node[main node] (3) [below right of=1] {$T_{swh}$};
          \node[main node] (4) [right of=3] {$Y_{swh}$};
          \node[main node] (5) [right of=1] {$X_{swh}$};
    
          \path[every node/.style={font=\sffamily\small}]
          (1) edge node  {} (3) 
          (1) edge node  {} (4) 
          (3) edge node  {} (4)
          (5) edge node  {} (3) 
          (5) edge node  {} (4); 
        \end{tikzpicture}
      \end{center}
      \vspace{2.5em}
      \caption{Baseline DAG. }\label{fig:SimpleDag}
  \end{subfigure}
  \hfill
  \begin{subfigure}{0.48\textwidth}
        \centering
        \tikzstyle{main node}=[circle,draw,font=\sffamily\small\bfseries]
        \tikzstyle{sub node}=[circle,draw,dashed,font=\sffamily\small\bfseries]
        \begin{tikzpicture}[->,>=stealth',shorten >=1pt,auto,node distance=2cm,thick]
          
          \node[sub node] (1) {$U_{swh}$};
          \node[main node] (2) [below of=1] {$I_{sh'w'}$};
          \node[main node] (3) [below right of=1] {$T_{swh}$};
          \node[main node] (4) [right of=3] {$Y_{swh}$};
          \node[main node] (5) [right of=1] {$X_{swh}$};
          
          \node[rectangle,draw=gray, fit=(2),inner sep=2mm,label=below:{$w',h'\in \Pi_s(hw)$}] {};

          \path[every node/.style={font=\sffamily\small}]
          (1) edge node  {} (3) 
          (1) edge node  {} (4) 
          (2) edge node  {} (1) 
          (3) edge node  {} (4)
          (5) edge node  {} (3) 
          (5) edge node  {} (4); 
        \end{tikzpicture}
        \caption{DAG with image-based confounding. }\label{fig:ConvDag}
    \end{subfigure}  
    \caption{Causal DAGs representing variables associated with a scene $s$. In our running example, $U_{swh}$  represents unobserved confounders, $X_{swh}$ observed confounders, $T_{swh}$ treatment and $Y_{swh}$ the outcome, all at location $w,h$ in scene $s$. In the right-hand DAG, latent confounders $U_{swg}$ are determined by a neighborhood $\Pi_s(hw)$ of the location $h, w$ in the image representing scene $s$.}
\end{figure}

As a baseline, consider the causal graph in  Figure~\ref{fig:SimpleDag}, which depicts a classical confounding relationship where a treatment of interest ($T_{swh}$), such as an anti-poverty intervention, is associated with factors (such as the presence of mineral extraction sites), that affect both the treatment and the outcome ($Y_{swh}$). Observed confounding variables are grouped in $X_{swh}$; unobserved confounders in $U_{swh}$.
Here, $s$ denotes the image scene (e.g., village), $w$ and $h$ denote horizontal and vertical location in that scene. It is important to emphasize that $s$ need not be spatially defined, as in the case of X-ray data, where the scene index could refer to a patient or body parts. 

\paragraph{Pixel-based Image Confounding}
We now turn to the causal model in Figure~\ref{fig:SimpleDag}, which depicts the kind of confounding dealt with in much of observational research \citep{rosenbaum1983central}. We extend this model to describe image-based confounding. First, we define this at the local level, where treatments are implemented at specific locations $h,w$ (in, for example, the precision agriculture context; \citet{liaghat2010review}). 
%
To give a precise example, we introduce the following notation. Let $\Pi_s(hw)\in \mathbb{N}^{2}$ denote the set of location indices involved in the spatial neighborhood analysis at $swh$. For example, in the case of a $z\times z$ square filter applied to the image ($z\in \mathbb{N}$), %
\begin{equation*}
{
\Pi_s(hw) = \big\{h-\lfloor z/2\rfloor,...,h+\lfloor z/2\rfloor\big\} \times 
\big\{w-\lfloor z/2\rfloor,...,w+\lfloor z/2\rfloor\big\}
}
\end{equation*}

With this notation, we can define the confounding structure induced by image-based confounding at the neighborhood level in Figure~\ref{fig:ConvDag}. Figure \ref{fig:ConvDag} is a formulation of spatial interdependence, as the image-information for indices in $\Pi_s(hw)$ affects the confounder $U_{swh}$. Conditional on the value of the confounder, the treatment decision for each unit made. This is illustrated by a decision maker who scans a scene looking for similarity of the neighborhood around $swh$ to some mental image, defined, for example, by an image filter $l$, and make a decision on this basis. 

%

\begin{figure}[t]
    \centering
    \begin{subfigure}[t]{.3\textwidth}
        \centering
        \includegraphics[width=.8\textwidth]{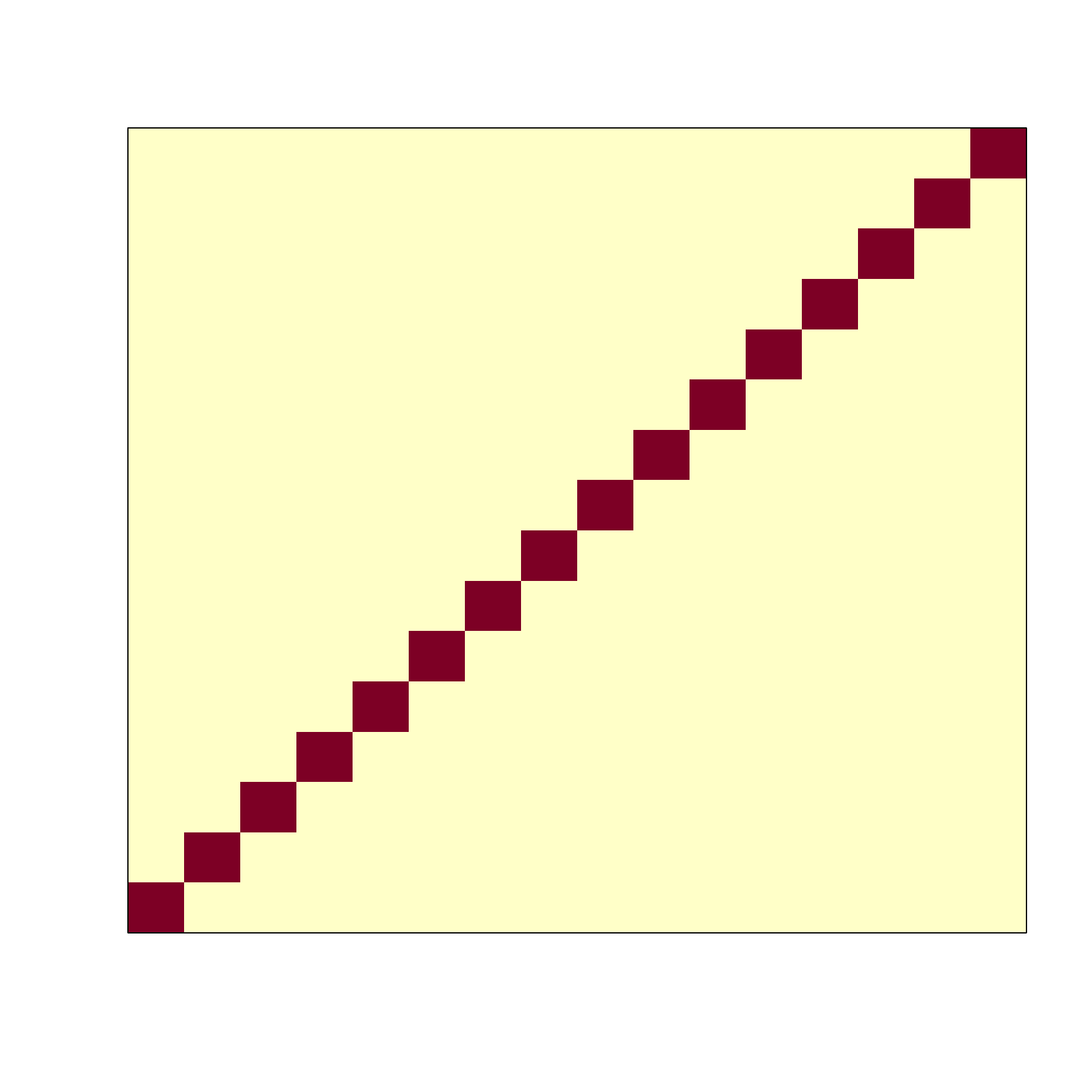}
        \vspace{-1em}
        \label{fig:convKern}
    \end{subfigure}
    \hfill
    \begin{subfigure}[t]{0.65\textwidth}
        \centering
         \includegraphics[width=.85\textwidth]{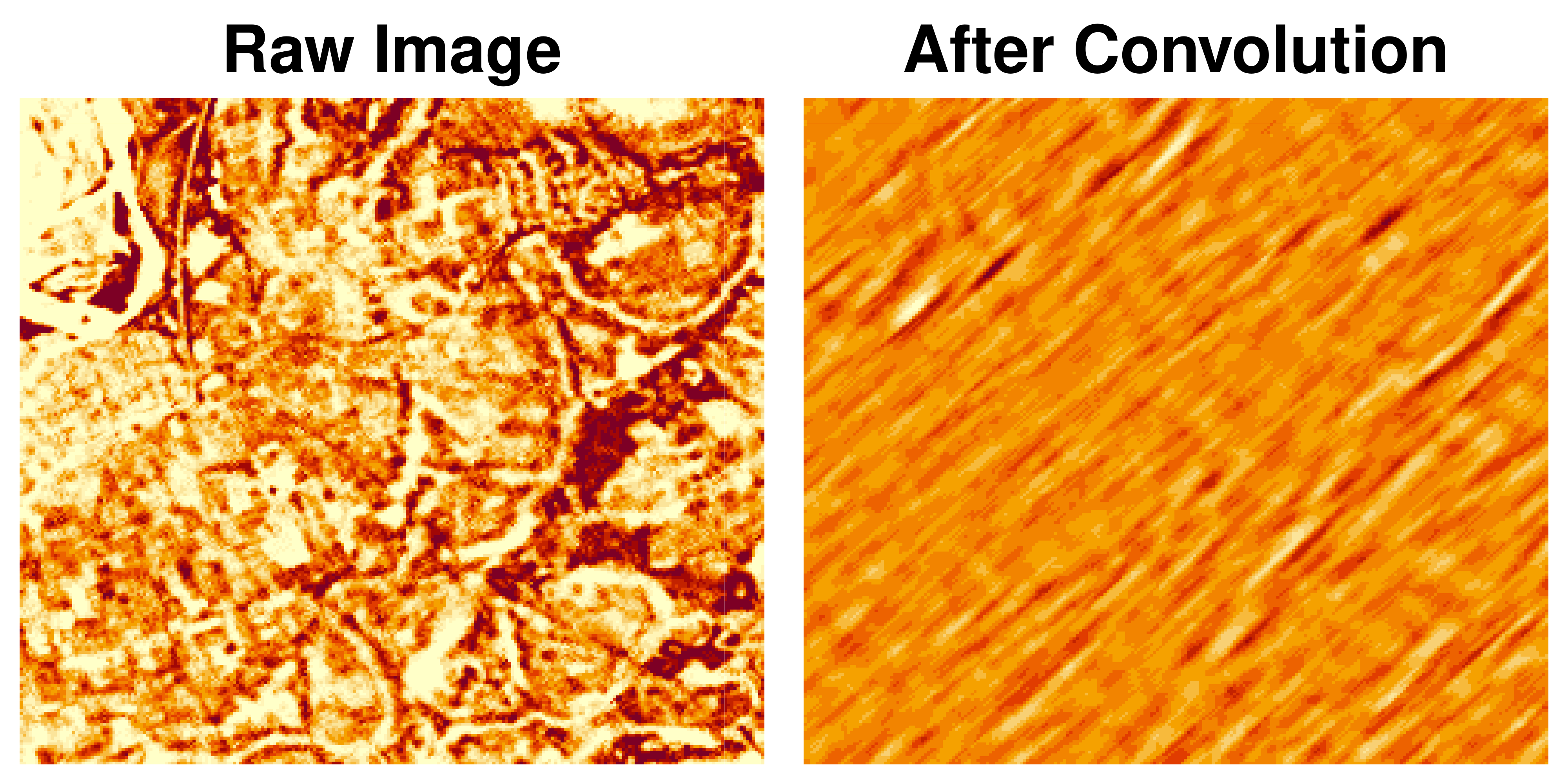}
    \end{subfigure}
        \caption{\emph{Left.} The kernel filter used to generate $U_{swh}$ in the right-hand image. \emph{Center, right.} Illustration of image-based confounding using Landsat data for Nigeria. The center panel depicts the raw reflectance; the right panel depicts the transformed values after convolution with the filter, values which enter the model for treatment/outcome to generate confounding.}\label{fig:conv}
\end{figure}

We illustrate this process on satellite image data from Landsat (see \S\ref{ss:data} for details). We let a single convolutional filter in the form of a diagonal matrix represent the image pattern pursued by a decision maker, as depicted in Figure \ref{fig:conv}. After applying $f_l$ to the raw image shown in the right panel of Figure \ref{fig:conv}, we obtain the resulting image-derived confounder values. This is a simple illustration of how the presence of objects in images (such as a diagonal line) can generate the confounder values.

\paragraph{Scene-based Image Confounding}
In practice, however, multiple scales are at play: the image is defined at one resolution, but the treatment, outcome, and confounder potentially at another. To take one example, the treatment and outcome data (e.g., mortality) are defined in the medical context at the patient level, but the X-ray data itself, of course, contains additional height and width dimensions. To take another example, a policymaker may examine an entire village, looking for the the best maximum or average similarity to some target pattern: the village is the unit to which treatment is allocated and the confounder is defined at that level, but is created using more granular information. Thus, our approach accommodates situations where the confounder, treatment and outcome are measured at different scales. For the sake of clarity, we will focus on the case where these variables measured at the same granularity, as this is one of the most common settings in applied research.  

Hence, we define the following causal model. Let $\Pi_s \in \mathbb{N}^{2}$ denote the height and width indices (locations) used when aggregating up information to the final scene-based unit of analysis, $s$. This scene-based causal model is significant because, while some studies are able to obtain resolved (e.g., household-level) outcome data,
this data may in other cases be costly or even impossible to obtain due to privacy reasons. In these situations, data can only measured at the scene level.

\begin{figure}[t]
\begin{subfigure}{.48\textwidth}
    \vspace{2.8em}
    \begin{center}
    \tikzstyle{main node}=[circle,draw,font=\sffamily\small\bfseries]
    \tikzstyle{sub node}=[circle,draw,dashed,font=\sffamily\small\bfseries]
    \begin{tikzpicture}[->,>=stealth',shorten >=1pt,auto,node distance=1.8cm,thick]
      
      \node[sub node] (0) {$U_{swh}$};
      \node[sub node] (1) [right of=0] {$U_s$};
      \node[main node](2)[below of =0, yshift=2mm]{$I_{sh'w'}$};
      \node[main node] (3) [below right of=1] {$T_s$};
      \node[main node] (4) [right of=3] {$Y_s$};
      \node[main node] (5) [right of=1] {$X_s$};
      
      \node[rectangle,draw=gray, fit=(0) (2),inner sep=7mm,label=below:{$w,h\in \Pi_{s}$}] {};
      
      \node[rectangle,draw=gray, fit=(2),inner sep=1mm,label=below:{\tiny $w',h'\in$ \newline $\Pi_s(hw)$}] {};

      \path[every node/.style={font=\sffamily\small}]
      (1) edge node  {} (3) 
      (1) edge node  {} (4) 
      (2) edge node  {} (0) 
      (0) edge node  {} (1) 
      (3) edge node  {} (4)
      (5) edge node  {} (3) 
      (5) edge node  {} (4) ;
    \end{tikzpicture}
  \end{center}
  \caption{Image-based confounding at the scene-level.}\label{fig:ComplexConvDag}
 \end{subfigure}
 \hfill
 \begin{subfigure}{.48\textwidth}
    \begin{center}
    \tikzstyle{main node}=[circle,draw,font=\sffamily\small\bfseries]
    \tikzstyle{sub node}=[circle,draw,dashed,font=\sffamily\small\bfseries]
    \begin{tikzpicture}[->,>=stealth',shorten >=1pt,auto,node distance=1.8cm,thick]
      
      \node[sub node] (0) {$U_{swh}$};
      \node[sub node] (1) [right of=0] {$U_s$};
      \node[sub node] (6) [above of=1] {$R_s$};
      \node[main node](2)[below of =0, yshift=2mm]{$I_{sh'w'}$};
      \node[main node] (3) [below right of=1] {$T_s$};
      \node[main node] (4) [right of=3] {$Y_s$};
      \node[main node] (5) [right of=1] {$X_s$};
      
      \node[rectangle,draw=gray, fit=(0) (2),inner sep=7mm,label=below:{$w,h\in \Pi_{s}$}] {};
      
      \node[rectangle,draw=gray, fit=(2),inner sep=1mm,label=below:{\tiny $w',h'\in$ \newline $\Pi_s(hw)$}] {};

      \path[every node/.style={font=\sffamily\small}]
      (1) edge node  {} (3) 
      (6) edge node  { Objects not in image} (1) 
      (1) edge node  {} (4) 
      (2) edge node  {} (0) 
      (0) edge node  {} (1) 
      (3) edge node  {} (4)
      (5) edge node  {} (3) 
      (5) edge node  {} (4);
    \end{tikzpicture}
  \end{center}
  \caption{Some confounding not observed in image.}\label{fig:ResidualBias}
  \end{subfigure}
  \caption{}
\end{figure}
 

\paragraph{SUTVA} Before discussing how we identify the causal effect of interest, we discuss how the Stable Unit Treatment Value Assumption (SUTVA) affects out problem formulation. This assumption entails that any unit \textit{i} treated should not  affect the treatment status our outcome of another unit \textit{j}. In other words, each unit is i.i.d, as defined by our DAG. However, in spatial analysis, it may be harder to defend SUTVA. Units that are closer in space may affect each other (via "spillover effects"). For example, a policymaker allocating aid to village \textit{i} may unintentionally affect aid in village \textit{j}. But SUTVA violations, and other form of dependence (e.g., spatial clustering), can in principle be accounted for by specifying an appropriate variance-covariance structure \citep{sinclairDetectingSpilloverEffects2012}. Still, we will focus on the simplest case where we assume SUTVA holds as a first approximation.

\paragraph{Identification}\label{ss:identification}
We confirm that, under image-based confounding as formalized in Figures \ref{fig:ConvDag} and \ref{fig:ComplexConvDag}, treatment effects may be identified by adjusting for the image $I$. To start, we assume that the confounder $U$ is a deterministic function of $I$ and return to a case where $U$ has multiple causes later. This is justified, for example, in applications where confounding is based on the existence of an object, either if the policymaker scanned $I$ for the object prompting the  policymaker to allocate a  treatment in that area of interest, or if $I$ can identify the object without error.
%
As $U$ is determined fully by $I$, ruling out other potential noise sources, there exists a deterministic function $f$ such that $U = f(I)$. The aforementioned case of $U$ being the (pooled) convolution of a 2D filter with the image $I$ satisfies this assumption. 
Here, we suppress dependence on the indexes $s,w,$ and $h$ since the same argument applies if $T/U/Y$ are defined at the $swh$ (pixel) or $s$ (scene) levels. 

\begin{proposition}\label{prop:identification}
Suppose the confounder $U$ is deterministic given the image $I$, such that $U = f(I)$, (with $f$ unknown), and that the causal structure obeys either of Figures \ref{fig:ConvDag} \& \ref{fig:ComplexConvDag}. Then, $p(Y(t))$ and therefore ATE of $T$ on $Y$ is identifiable from $p(I, X, T, Y)$.
\end{proposition}
\begin{proof} For simplicity of exposition, we give the proof for the case without additional  confounding variables $X$. The proof generalizes readily to non-empty $X$, by marginalization and conditioning. The claim follows from $U$ being a deterministic function of $I$. By the backdoor criterion applied to the graphs in Figures \ref{fig:ConvDag} \& \ref{fig:ComplexConvDag}, $X, U$ is an adjustment set for the effect of $T$ on $Y$, which implies exchangeability of potential outcomes: $Y(t) \perp T \mid X$, see e.g., \citet{hernanbook}. In the case with empty $X$,
\[
p(Y(t)) = \sum_{u} p(Y|T=t,U=u)p(U=u).
\]
Since $U$ is a deterministic, but not necessarily invertible function of $I$, $U=f(I)$, we have that 
\begin{align}%
p(Y\mid T=t, U=u) &= p(Y\mid T=t, I \in f^{-1}(u)) \;\;\mbox{ and }\;\;
p(U=u) &= \sum_{i \in f^{-1}(u) } p(I=i) \label{eq:marginal-u}
\end{align}%
where $f^{-1}$ is the inverse map of $f$, so that 
\begin{align*} 
p(Y(t)) &= \sum_u \sum_{i \in f^{-1}(u) } p(Y|T=t,I=i)p(I=i) = \sum_i p(Y|T=t,I=i)p(I=i)~.
\end{align*} 
Hence, $I$ is also an adjustment set for $T$ on $Y$. From a similar proof, we see that $X, I$ is an adjustment set in the case of non-empty $X$. From here, standard arguments \citep{rosenbaum1983central} show that
$$
\mathbb{E}[Y(t)] = \mathbb{E}\left[Y \frac{p(T=t)}{p(T=t\mid I=i)} \ \bigg| \ T=t \right]
$$
which justifies use of inverse-propensity weighting with respect to $I$.
\end{proof}

\paragraph{Image an Imperfect Confounder Proxy}\label{ss:imperfect}
The argument of Proposition~\ref{prop:identification} rests on the assumption that the image contains all information about the latent confounder. When treatment decisions are made based on object detection, this assumption would be satisfied if the image contains all objects that are relevant to the outcome and treatment decision. This is violated if, for example, unlabeled objects, depicted as $R_s$ in Figure \ref{fig:ResidualBias}, that are themselves the driver of the treatment decision but are not possible to reconstruct from the image data. If the image data imperfectly depicts those objects, full identification is no longer possible as there is a possibility of residual confounding. Specifically, the inverse map $f^{-1}$ in \eqref{eq:marginal-u} is no longer uniquely or well defined as a set of images; $R_s$ must be adjusted for as well. In this imperfect case, the image becomes a \textit{driver} of the confounding, and thus, has similar properties to proxies \citep{pearlLinearModelsUseful2013,penaMonotonicityNondifferentiallyMismeasured2020a}.

\paragraph{Central Role of Resolution} For satellite imagery data of particular relevance to social and health science applications, resolution is a key driver of this residual confounding. We can only adjust for confounder objects in the image that can be resolved. Smaller confounder objects therefore introduce residual bias, indicating how technological improvements to sensor technology play a critical role in improving image-based causal inference methods. 



\paragraph{Finite Sample Estimation Considerations}\label{ss:estimation}
We have shown how, under assumptions, the image is itself an adjustment set for estimating the effect of programs on outcomes in the context of image-based confounding. However, non-parametric inference is difficult in the image context because no two images are the same. Thus, the probability of seeing comparable treatment/control images is zero, violating overlap assumptions necessary for model-free inference  \citep{d2019comment}. Machine learning models for the image may seek to estimate $U$, forming latent representations for the image. In this lower-dimensional space, there is more likely to be empirical overlap between treatment/control, justifying the use of modeling approaches like the ones discussed in \S\ref{ss:experiments} and \S\ref{sec:mu_empirical}. Thus, while adjusting directly for $U$ would fulfill the overlap assumptions optimally, this is infeasible; when adjusting for $I$ instead, a critical argument for our approach to work is that the propensities depend only on aspects of $I$ that captures $U$, aspects assumed to be compressible to a lower dimensional representation. As such, the situation is more benign than if propensities depended freely on all patterns in $I$.




\section{Experiments Illustrating Performance Under Model Misspecification}\label{ss:experiments}
Although the identification results described in \S\ref{ss:identification} are general, they are also theoretical and, as described in \S\ref{ss:estimation}, there are several practical considerations that will affect performance in real data. In order to understand these dynamics, we use simulation, since (in practice) true causal targets are unknown. In these simulations, the image data is observed, but the confounding features are not, and must be estimated via machine learning, as is the case in real applications. 

\paragraph{Simulation Design} In first simulation, pixel-level confounding is generated via convolution applied to our set of Landsat images from Nigeria:
\[
U_{swh} = f_l(I_{s\Pi_s(hw)}),
\]
where $f_l(\cdot)$ denotes the linear kernel function, where the single kernel filter, $l$, is a diagonal matrix and is shown in Figure \ref{fig:conv}. The set of images used in the simulation are drawn from the corpus of Landsat satellite images that we later use in the application (see \S\ref{ss:data}). 

Because the confounder, $U_{swh}$ enters the equation both for the treatment probability and the outcome, we have confounding: 
\begin{align*} 
\Pr(T_{swh} | I_s) &= \textrm{logistic}( \beta  U_{swh} + \epsilon_{swh}^{(W)}) \;\;\mbox{ and }\;\;  Y_{swh} = \gamma   U_{swh} + \tau \, T_{swh} + \epsilon_{swh}^{(Y)},
\end{align*} 
where the error terms, $\epsilon_{swh}^{(W)}$ and $\epsilon_{swh}^{(Y)}$, are drawn i.i.d. from $N(0,0.1)$ and $\textrm{logistic}(x)=1/(1+\exp(-x))$ maps real numbers to values between 0 and 1. 
We vary the dimensions of the convolutional filter, $l$, keeping the structure of the filter pattern fixed (see Figure \ref{fig:conv}) but normalizing as we vary the dimensionality so that the convolution output values have the same mean and variance. By varying the filter dimensions, we alter the size of the neighborhood used in calculating the pixel-level confounder. A larger filter size means that the treatment allocator scans a larger region around each point in assessing proximity to the target pattern. 

\paragraph{Estimators} 
We compare three estimators. First, we examine the difference-in-means estimator, $\hat{\tau}_{\textrm{0}}$, defined as the difference in conditional outcome means for the two treatment groups. Because of confounding, this quantity is biased for $\tau$. We use this value as a baseline and report our evaluation metrics relative to those from this estimator in order to account for the fact that, as we vary parameters in the data-generating process, the natural scale of the estimates varies as well. 

We also report results from two Inverse Propensity Weighting (IPW) estimators \citep{austin2015moving}, one using the true (``oracle'') propensities that are not known in practice and another using a single-layer ConvNet which is trained using backpropogation with the binary cross-entropy loss. The estimation formula for IPW is $\hat{\pi}(\bX_s)=\widehat{\textrm{Pr}}(T_s=1|I_s)$, $\widehat{\tau} =  \frac{1}{n}\sum_{s=1}^n \left\{ \frac{T_s Y_s}{\hat\pi(\bX_s)} -\frac{(1-T_s)Y_s}{1-\hat\pi(\bX_s)}\right\}$ for the scene analysis and is defined analogously at the pixel level. We report results from the Hajek IPW estimator where the weights have been normalized, reducing estimator variance at the cost of some finite sample bias \citep{skinner2017introduction}. 

\paragraph{Evaluation Metrics} 
For our evaluation, we compare the true $\tau$ with the estimated values. We report the absolute bias of $\hat{\tau}$ and the Root Mean Squared Error (RMSE): 
\begin{equation*}
\textrm{Absolute Bias} = \big|\E[\hat{\tau} - \tau]\big|; \;\;
\textrm{RMSE} = \sqrt{\E[(\hat{\tau} - \tau)^2 ]},
\end{equation*}
where expectations are taken over the data-generating process and are approximated via Monte Carlo. As noted, these evaluation metrics are then normalized using those from the baseline $\hat{\tau}_0$.

\begin{figure}[t]
    \centering
        \centering
        \includegraphics[width=0.52\linewidth]{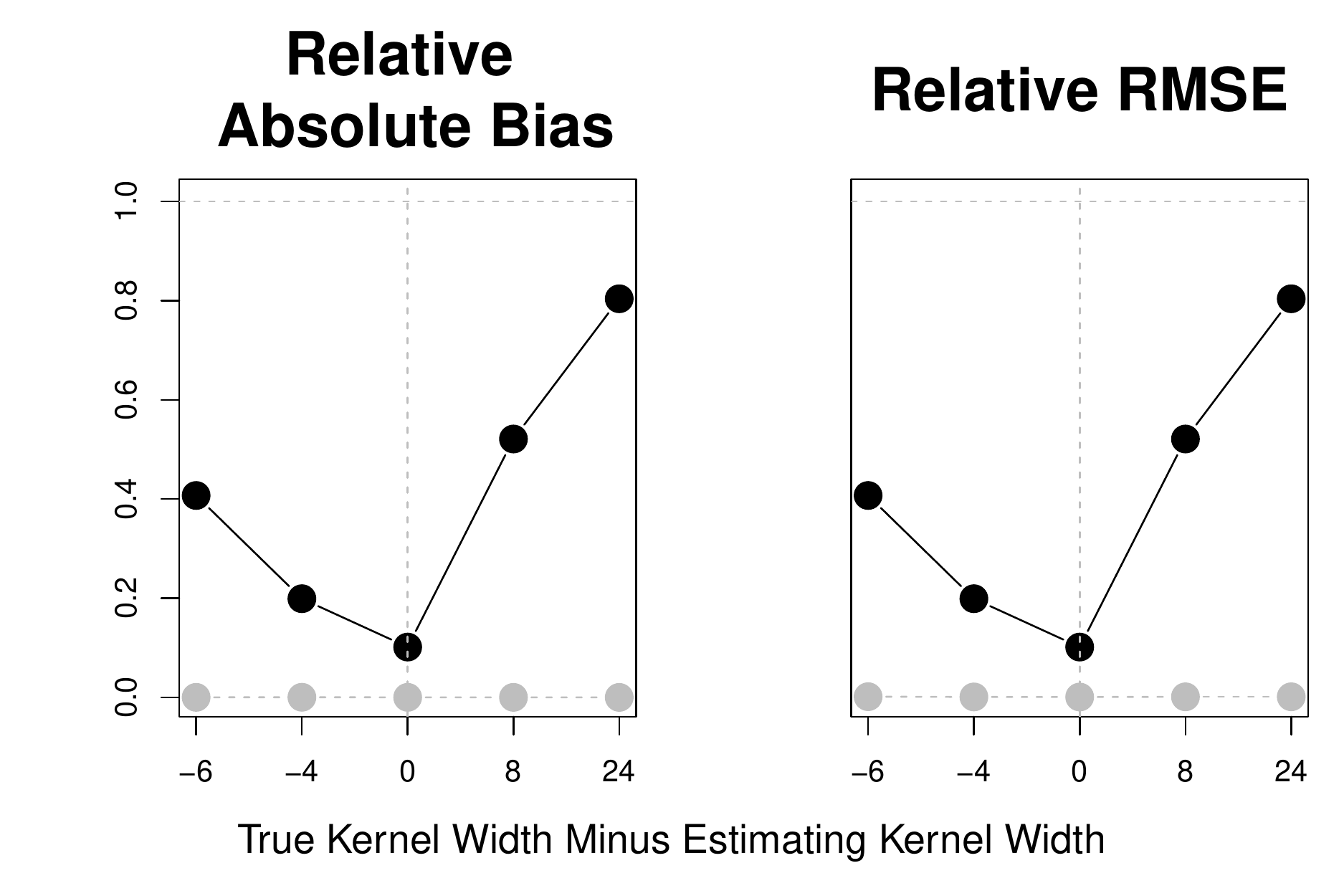}
        \caption{Pixel-level results varying extent of the spatial confounding. Results indicated with black circles are obtained using the estimated propensities; results indicated with gray circles are obtained using the true (oracle) values. The estimating kernel width is fixed at 8; the true kernel width varies.}\label{fig:PixelResultsVaryGlobality}
\end{figure}

\paragraph{Simulation Results}
First, in Figure~\ref{fig:PixelResultsVaryGlobality}, we show performance where the treatment and outcome are defined at the pixel-level. The bias and RMSE panels of Figure \ref{fig:PixelResultsVaryGlobality} look similar because the variance of estimation is small due to the large number of pixel realizations from which to estimate parameters. 

As expected, we also find that the absolute bias and RMSE are minimized when the kernel width used in estimation is the same as the kernel width used to generate the true confounder. The bias and RMSE grow larger when the true kernel width is greater than the estimating kernel width. This finding is likely due to the fact that, when the neighborhood size used in estimation is larger than that used to create the confounder, the extra, irrelevant information can be ignored by the model. Conversely, when the neighborhood size generating the true confounding is larger than what the model can capture, bias is more severe because the model cannot account for this extra information. We find similar results for the scene-level simulations (see \ref{ss:AdditionalSims}).  

\section{Empirical Illustration: The Impact of Local Aid Programs in Nigeria}\label{sec:mu_empirical}
Having examined finite-sample performance via simulation, in this section, we demonstrate our method in Nigeria, African's largest economy and a country that is projected to be the world's second most populous by 2100 \citep{vollset2020fertility}. Despite an average economic growth rate of around 3\% since 2000, about 40 percent of the Nigerian population live below the poverty line (\$2 per day). In response, governments and NGOs have deployed a variety of local aid programs to the country. However, the causal impact of these programs is difficult to estimate \citep{roodman2008through}. 

While geo-temporal data on poverty, $Y$, and interventions, $T$, are readily available, there is a lack of geo-temporal data on potential confounders, $U$, at the local level. While some of these confounders may be difficult to capture directly using images (such as the quality of political institutions), there may be information present in remote sensing imagery about other confounder objects related to infrastructure or agriculture \citep{schnebele2015review,steven2013applications}. We therefore use satellite images of Nigerian communities in order to estimate the impact of local aid programs. 

\paragraph{Data Description}\label{ss:data}
Our outcome data on poverty is drawn from the Demographic and Health Surveys (DHS), which are conducted by a non-profit organization, ICF International, with funding from USAID, WHO, and other international organizations \citep{rutstein2006guide}. The DHS surveys are conducted at the household level for randomly selected clusters that aggregate to geographically explicit scenes of about 300 households for de-identification purposes. Our outcome measure is the International Wealth Index (IWI) from 2018, which is a principal-components-derived summary of 12 variables including households' access to public services and possession of consumer products such as a phone. Its scale is between 0 and 100, with higher values indicating greater wealth. 

Our treatment data is drawn from a data set on international aid programs to Nigeria compiled by AidData \citep{aiddata2016} used under an Open Data Commons License. The aid programs we examine took place after 2003. The aid providers include entities such as the World Bank and WHO, as well as domestic governments such as the United States. The programs we examine are diverse, focusing on infrastructure (e.g., support for road development) and agriculture (e.g., support for small-scale farmers), among other things. For the simplicity of our presentation, we take the presence of a geographically-specific aid program within 7000 meters of a DHS point as our treatment. 

\begin{figure}[t]
\begin{subfigure}{.52\textwidth}
    \centering
    \includegraphics[width=1.\textwidth]{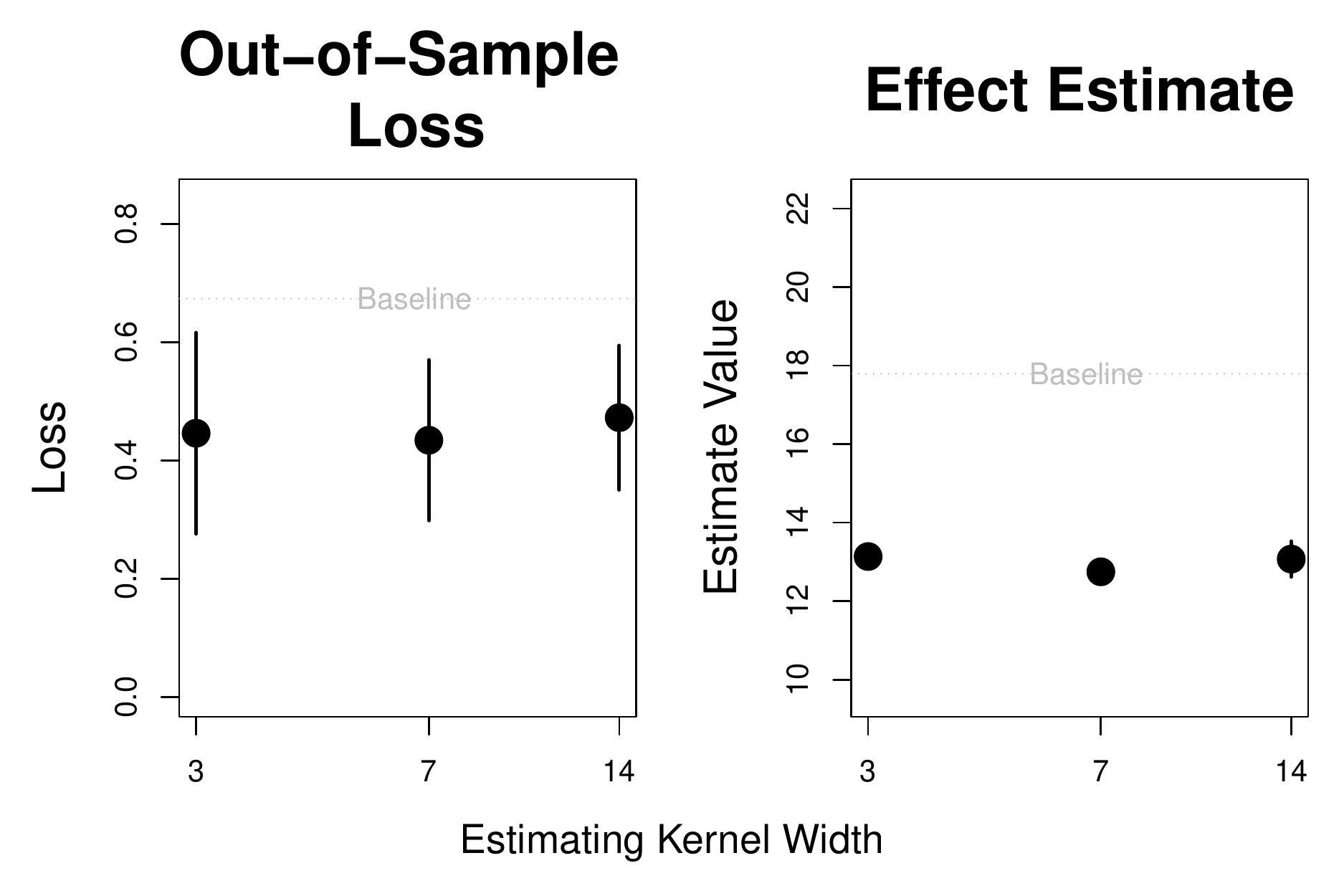}
    \caption{The left panel shows out-of-sample binary cross entropy loss compared to the baseline loss when guessing the dominant class. The right panel shows the IPW estimate values across the range of estimating kernel width values compared to the baseline difference in conditional means.}\label{fig:ApplicationResults}  
\end{subfigure}
\hfill
\begin{subfigure}{.46\textwidth}
    \centering
    \includegraphics[width=.7\textwidth]{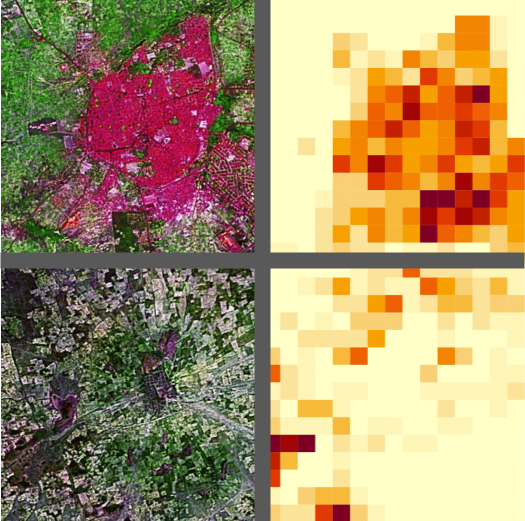}
    \caption{ The left two panels depict the raw data for a treatment (top) and control (bottom) DHS point. The right two panels show the final convolutional layer for each case.}\label{fig:InsideConvnet}
\end{subfigure}
\caption{}
\end{figure}

Our pre-treatment image data is drawn from Landsat, a satellite imagery program operated by NASA/USGS. We use the Orthorectified ETM+ pan-sharpened data derived from the raw satellite imagery captured between 1998 and 2001; the raw data have been processed to contain minimal cloud-cover and to be correctly geo-referenced. Resolution is 14.25 meters; reflectance in the green, near-infrared, and short-wave infrared bands is recorded. Example images of treated/control DHS points are shown in the left two panels of Figure \ref{fig:InsideConvnet} (top/bottom, respectively).

\paragraph{Model Design}\label{ss:model}

Our image model for the treatment is built using three convolutional layers (32 filters each) with max pooling. After the application of the filters, we project the channel dimension into a one-dimensional space to facilitate interpretability via a single image post-convolution. Batch normalization is used on the channel dimensions and before the final projection layer. Weights are learned using ADAM with Nesterov momentum with cosine learning rate decay (baseline rate = 0.005; \citet{gotmare2018closer}). We compare performance across a variety of filter sizes. We attempt to limit overfitting by randomly reflecting each image dimension with probability $1/2$ during training. We assess out-of-sample performance using three random training/test splits (1277/20), averaging over this sampling process and reporting 95\% confidence intervals from the three test set assessments.

\paragraph{Empirical Results}\label{ss:ApplicationResults}
First, in the left panel of Figure \ref{fig:ApplicationResults}, we assess propensity model performance. We find that the image model always improves on the baseline out-of-sample loss value obtained by random guessing of the dominant class (the control class, comprising 71\% of the data sample). Performance is stable across values of the estimating kernel width. 

Next, in the right panel of Figure \ref{fig:ApplicationResults}, we analyze the estimated treatment effects. We find that, across the estimating kernel width range, adjusted estimates are positive but smaller in magnitude than the baseline difference in means. This hints at the importance of confounder adjustment, as these programs may be given to areas already primed for growth. 

We also analyze the model output. In Figure \ref{fig:InsideConvnet}, we visualize data for the out-of-sample treated unit with the highest treatment probability (top) and control unit with the lowest probability (bottom). The left panels show the raw data and the right panels show the final hidden convolutional layer pre-flattening. The low treatment probability site is in the remote desert city of Machina (pop. 62,000); the high probability site is from the city of Katsina (pop. 429,000) near a large agricultural basin. The output of this model would seem to resonate with the fact that many of the projects undertaken by global actors are specifically designed to assist farmers and agriculture more generally. 

Finally, we access the performance of the estimated propensity scores on reducing covariate imbalance between treated and control groups. The same absence of rich covariate information in places such as sub-Saharan Africa that motivates this paper also makes this assessment task difficult. Still, we can analyze differences in longitude/latitude between treated/control groups. We find a raw difference of (-1.17, -1.13). After weighting, this difference decreases in magnitude (e.g., to (-0.63,-0.30) in a representative run), indicative of improved counterfactual comparisons. 

\section{Discussion and Future Work}\label{s:discussion}
In this paper, we show how investigators can use machine-learning innovations for image prediction to adjust for confounding due to latent objects present in images - objects that affect both outcome and treatment decisions. We formally show that, even though these objects are latent, we can adjust for them using the image information alone. We illustrate this approach in order to understand the causal effect of aid programs on reducing poverty in Africa's largest but still poverty-affected economy. 

Our approach has some limitations. For example, as described in \S\ref{ss:imperfect},  confounding may be due to objects that cannot be resolved in the image data, and, as a result bias reduction will not occur conditioning on the image information. In this context, the collection and application of imagery with higher spatial, temporal, and spectral resolution is a priority. Spatial and temporal resolution may both be achieved, for example, using imagers mounted on ground-based infrastructure (e.g., \citet{johnston2021measuring}); spatial resolution and extent could be optimized with drone- or airplane-based instruments (e.g., \citet{gray2018integrating}). Future considerations should examine privacy and fairness issues with causal analyses based on passive sensor technologies. 

Second, while in some cases, such as healthcare, the scene-level unit of analysis is clearly defined (e.g., the patient), in other context, the scene-level unit of analysis is more ambiguous. The researcher therefore has choices about how to define the scene (e.g., at the street, village, or region levels), a choice that could introduce systematic bias into the analysis. This issue is known as the modifiable areal unit problem \citep{fotheringham1991modifiable}, and the approach described here is vulnerable to it as well. Future work should therefore focus on the development of image-confounding methods that have theoretical guarantees on the robustness of the results to the scale of action examined. This issue of scale is also related to the question of capturing the treatment and outcome at different scales.

Third, our model learns the filters that best predict treatment assignment. More research is needed to connect these patterns, learned inductive, with the mental processes of real actors as they consult images in decisionmaking. This path of research could forge interresting links between cognitive science, machine learning, and causal inference.  \hfill $\square$


\medskip

\bibliographystyle{plainnat}
\bibliography{confoundingbib}

\appendix

\renewcommand{\thefigure}{A.\arabic{figure}}
\setcounter{figure}{0}

\section{Appendix}
\subsection{Additional Simulation Results}\label{ss:AdditionalSims}
\paragraph{Scene-level Unit of Analysis}
Now, define the treatment and outcome at the scene level. We again have the convolutional structure in generating $U_{swh}$, by apply a max pooling function on the pixel-level confounders: 
\begin{align*} 
U_{s} &= \textrm{GN}(\max\{ U_{swh}: {hw \in \Pi_{s}} \}),
\\ \Pr(T_{s} | I_s) &= \textrm{logistic}( \beta  U_{s} + \epsilon_{s}^{(W)})
\\ 
Y_{s} &= \gamma    U_{s} + \tau \, T_{s} + \epsilon_{s}^{(Y)}, 
\end{align*} 
where GN($\cdot$) denotes normalization to mean 0 and variance 1, done to prevent all units receiving treated. We again vary the size of the convolutional filter, altering the spatial extent of confounding. 

\begin{figure}[h]
    \centering
        \centering
        \includegraphics[width=0.75\linewidth]{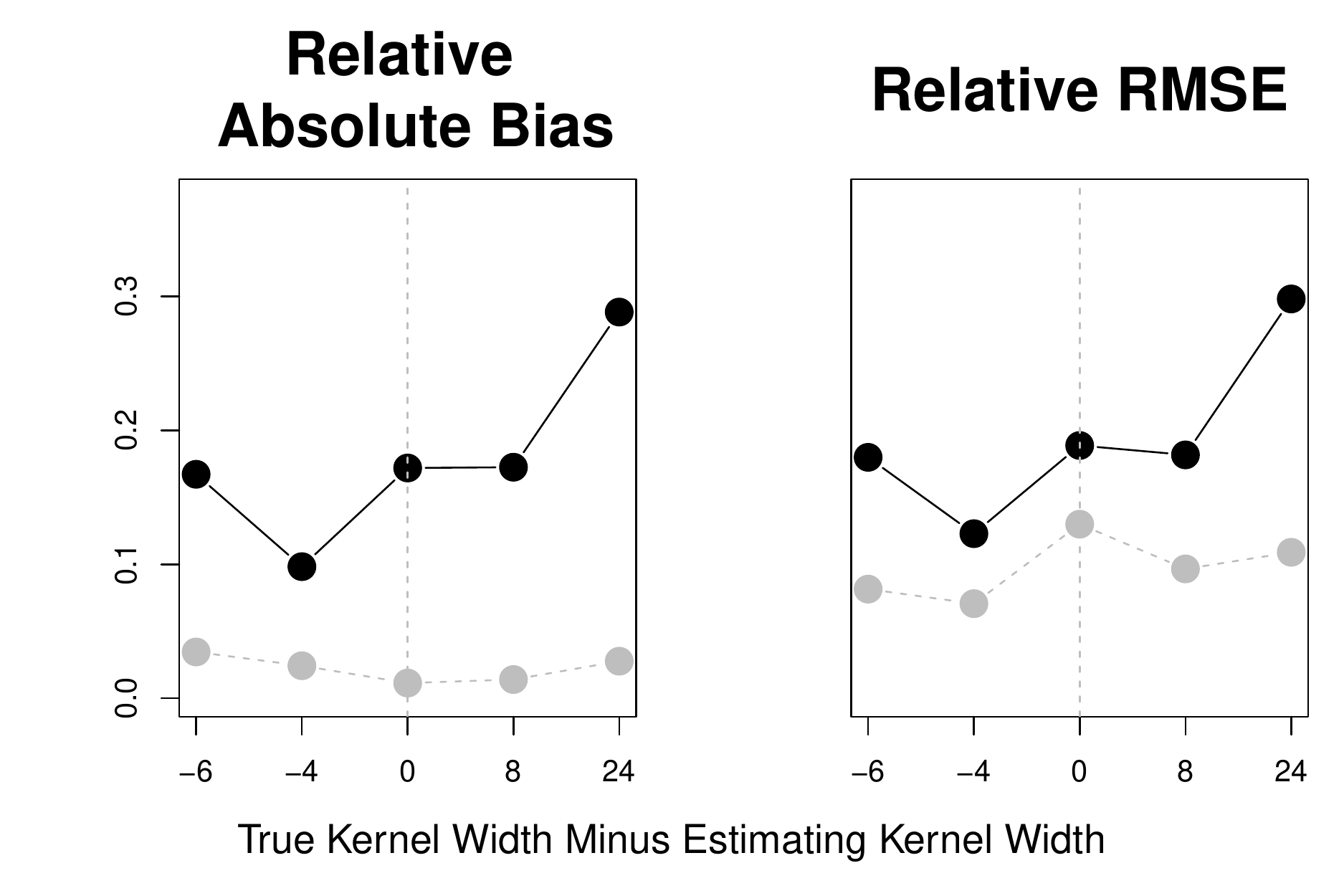}
        \caption{Bias and RMSE for the scene-level analysis as we vary the geographical size of the spatial confounding, holding the estimation kernel width fixed. The estimating kernel width is fixed at 8 and the true kernel width varies. Gray circles indicate  values using the true (oracle) propensities. }\label{fig:SceneResultsVaryGlobality}
\end{figure}

\paragraph{Scene-level Results}
In Figure \ref{fig:SceneResultsVaryGlobality}, we see similar dynamics as in the pixel-level analysis. We again find the same general pattern where overestimating the neighborhood size induces less bias than underestimating the neighborhood size. However, the scene-level bias seems to be somewhat less sensitive to the choice of estimating kernel width compared to the pixel-level analysis, where the application of the kernel filter is the only step in generating the latent confounder. The estimation and Monte Carlo variance are somewhat greater in this scene-level analysis because, with fewer draws of $W_s$ compared to $W_{swh}$, there is considerably less information to use when estimating parameters. 

\subsubsection{Probing Estimation Bias as the Determinism Assumption is Relaxed}
We have just explored how model misspecification affects estimation error. We now probe how relaxing the determinism assumption of Proposition \ref{prop:identification} affects performance. In particular, we now let the unobserved confounder be a random function of the image, as depicted visually in Figure \ref{fig:ResidualBias}. In particular, the confounder values are now 
\[
U_{swh} = f_l(I_{s\Pi_s(hw)}) + \epsilon_{shw}^{(U)},
\]
where $\epsilon_{shw}^{(U)}
\sim \mathcal{N}(0,\sigma_U^2)$. We vary $\sigma_U^2 \in \{1,3,5,7\}$. We then apply the same data generating process to obtain scene-level treatments and outcomes. 

\begin{figure}[h]
    \centering
        \centering
        \includegraphics[width=0.75\linewidth]{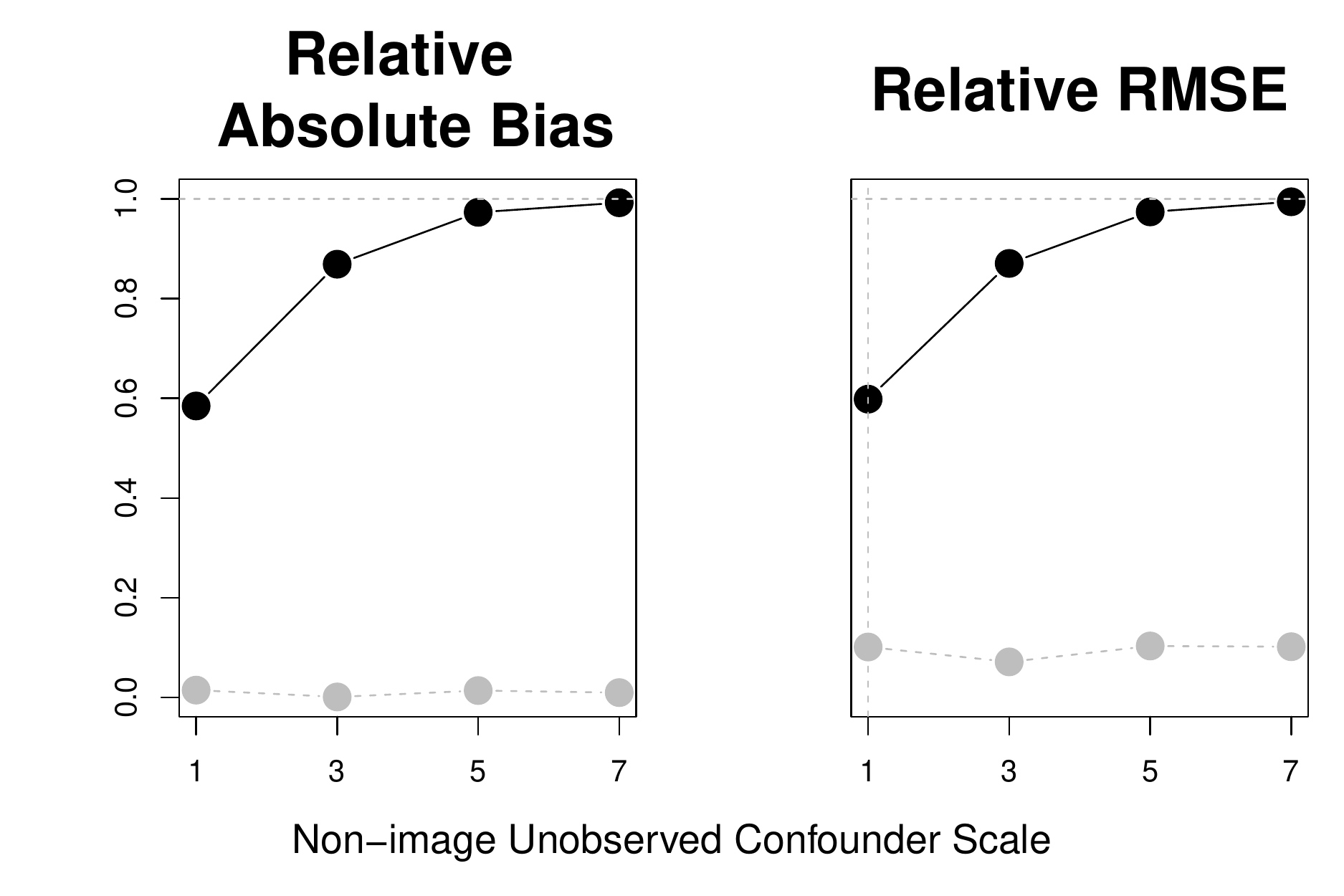}
        \caption{Bias and RMSE for the scene-level analysis as we vary the stochasticity present in the confounding mechanism, holding the true and estimation kernel width fixed at 8. Gray circles indicate  values using the true (oracle) propensities. }\label{fig:SceneResultsVaryGlobalityNoise}
\end{figure}

We see in Figure \ref{fig:SceneResultsVaryGlobalityNoise} how performance is affected relaxing determinism. As expected, we see that estimation bias grows as the unobserved confounding is increasingly determined by the noise factor, $\epsilon_{shw}^{(U)}$. When the noise scale is at its maximum, bias is still no worse than the simple difference in means baseline (i.e. relative bias/RMSE approaches 1). This fact is likely because the noise injected into the confounding mechanism is itself exogenous. Nevertheless, having established a theoretical baseline in this paper, future research should examine this noise-induced confounding to image-based causal inference in greater detail, connecting this line of work with the proxy literature.

%

\subsection{Implementation Details}
We implement analyses on an Apple M1 GPU using Metal-optimized tensorflow 2.8. Total compute time for the simulations is about 8 hours; total compute time for the application results is about 24 hours. 

\end{document}